\documentclass[]{bytedance_seed}

\usepackage[toc,page,header]{appendix}

\usepackage{minitoc}
\usepackage{amsthm}
\allowdisplaybreaks
\usepackage{mdframed}
\usepackage{booktabs}
\usepackage{array}
\usepackage{multirow}
\usepackage{graphicx}
\usepackage{tabularx}
\usepackage{multirow}
\usepackage{booktabs}
\usepackage{ragged2e}
\usepackage{xltabular}
\usepackage{longtable}
\usepackage{wrapfig}
\definecolor{darkgrey}{rgb}{0.53,0.53,0.53}
\definecolor{middlegrey}{rgb}{0.53,0.53,0.53}
\definecolor{mygrey}{rgb}{0.9,0.9,0.9}
\definecolor{mydarkblue}{rgb}{0,0.08,0.45}
\definecolor{darkdarkblue}{rgb}{0.0,0.0,0.3}
\definecolor{darkblue}{rgb}{0.0,0.0,0.7}
\definecolor{darkred}{rgb}{0.4,0,0.3}
\definecolor{lightblue}{HTML}{EDE6DC}
\definecolor{lightred}{HTML}{FFFAFA}
\definecolor{fancyblue}{HTML}{4771E3}
\definecolor{grey}{rgb}{0.95,0.95,0.95}
\definecolor{seedblue}{RGB}{46,90,168}
\definecolor{ForestGreen}{RGB}{34,139,34}

\title{
ProtoReasoning: Prototypes as the Foundation for Generalizable Reasoning in LLMs
}

\author[1,\dagger]{Feng He}
\author[1,2,*]{Zijun Chen}
\author[1]{Xinnian Liang}
\author[1]{Tingting Ma} 
\author[1]{Yunqi Qiu} 
\author[1, \dagger]{\\ Shuangzhi Wu}
\author[2]{Junchi Yan}

\affiliation[1]{ByteDance Seed}
\affiliation[2]{Shanghai Jiao Tong University}

\contribution[*]{Work done at ByteDance Seed}
\contribution[\dagger]{Corresponding authors}

\abstract{
\begin{abstract}

Recent advances in Large Reasoning Models (LRMs) trained with Long Chain-of-Thought (Long CoT) reasoning have demonstrated remarkable cross-domain generalization capabilities. However, the underlying mechanisms supporting such transfer remain poorly understood. 
We hypothesize that cross-domain generalization arises from \textbf{shared abstract reasoning prototypes} --- fundamental reasoning patterns that capture the essence of problems across domains. 
These prototypes minimize the nuances of the representation, revealing that seemingly diverse tasks are grounded in shared reasoning structures.
Based on this hypothesis, we propose ProtoReasoning, a framework that enhances the reasoning ability of LLMs by leveraging scalable and verifiable prototypical representations (Prolog for logical reasoning, PDDL for planning).
ProtoReasoning features: (1) an \textbf{automated} prototype construction pipeline that transforms problems into corresponding prototype representations; (2) a comprehensive \textbf{verification} system providing reliable feedback through Prolog/PDDL interpreters; (3) the \textbf{scalability} to synthesize problems arbitrarily within prototype space while ensuring correctness. 
Extensive experiments show that ProtoReasoning achieves 4.7\% improvement over baseline models on logical reasoning (Enigmata-Eval), 6.3\% improvement on planning tasks, 4.0\% improvement on general reasoning (MMLU) and 1.0\% on mathematics (AIME24). 
Significantly, our ablation studies confirm that learning in prototype space also demonstrates enhanced generalization to structurally similar problems compared to training solely on natural language representations, validating our hypothesis that reasoning prototypes serve as the foundation for generalizable reasoning in large language models.
\end{abstract}
}

\date{\today}
\correspondence{Feng He at \email{hefeng.hhh@bytedance.com}}

\begin{document}
\maketitle


\section{Introduction}

Large Reasoning Models (LRMs) trained through Reinforcement Learning with Verifiable Rewards (RLVR) have demonstrated remarkable capabilities in complex reasoning tasks~\citep{guo2025deepseek, team2025kimi, seed2025seed-thinking}. 
An intriguing phenomenon observed in recent research is that models trained with Long Chain-of-Thought (Long CoT) reasoning~\citep{chen2025longcot-survey} in one domain exhibit significant generalization abilities in other domains.
Notably, DeepSeek-R1~\citep{guo2025deepseek} generalizes from mathematical and coding domains to STEM and creative writing, while Logic-RL~\citep{xie2025logic-rl} transfers logical puzzle-solving capabilities to mathematical reasoning.

Despite these empirical successes, the underlying mechanisms enabling cross-domain generalization remain poorly understood.
A key question is: what allows models trained on specific reasoning tasks to transfer their abilities to different types of problems?
We hypothesize that this transfer capability stems from the existence of abstract \textbf{reasoning prototypes} --- fundamental reasoning patterns and thinking structures that underlie diverse problem domains~\cite {yang2025reasonflux}.
Specifically, when two problems share the same or similar reasoning structures in their underlying solution processes, regardless of their apparent differences in representation, strong generalizations will emerge between them.

\begin{figure*}[!tb]
    \centering
    \includegraphics[width=0.95\textwidth]{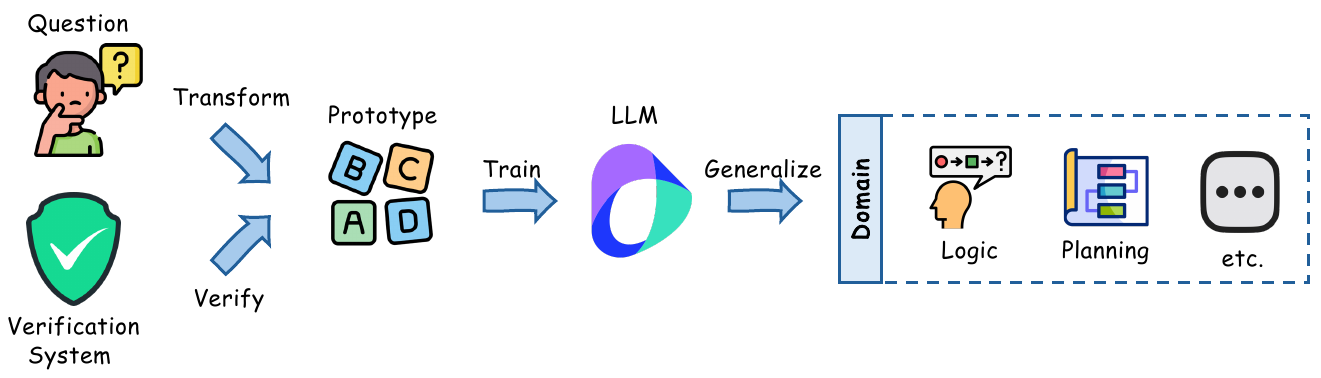}
    \caption{
    \textbf{ProtoReasoning Overview}: Our ProtoReasoning framework adopts prototypes as core representations to enhance model reasoning capabilities. The verification system built upon prototype systems provides accurate supervisory signals, ultimately enabling effective reasoning generalization.
    }
    \label{fig:overview}
\end{figure*}

\begin{figure*}[!tb]
    \centering
    \includegraphics[width=0.75\textwidth]{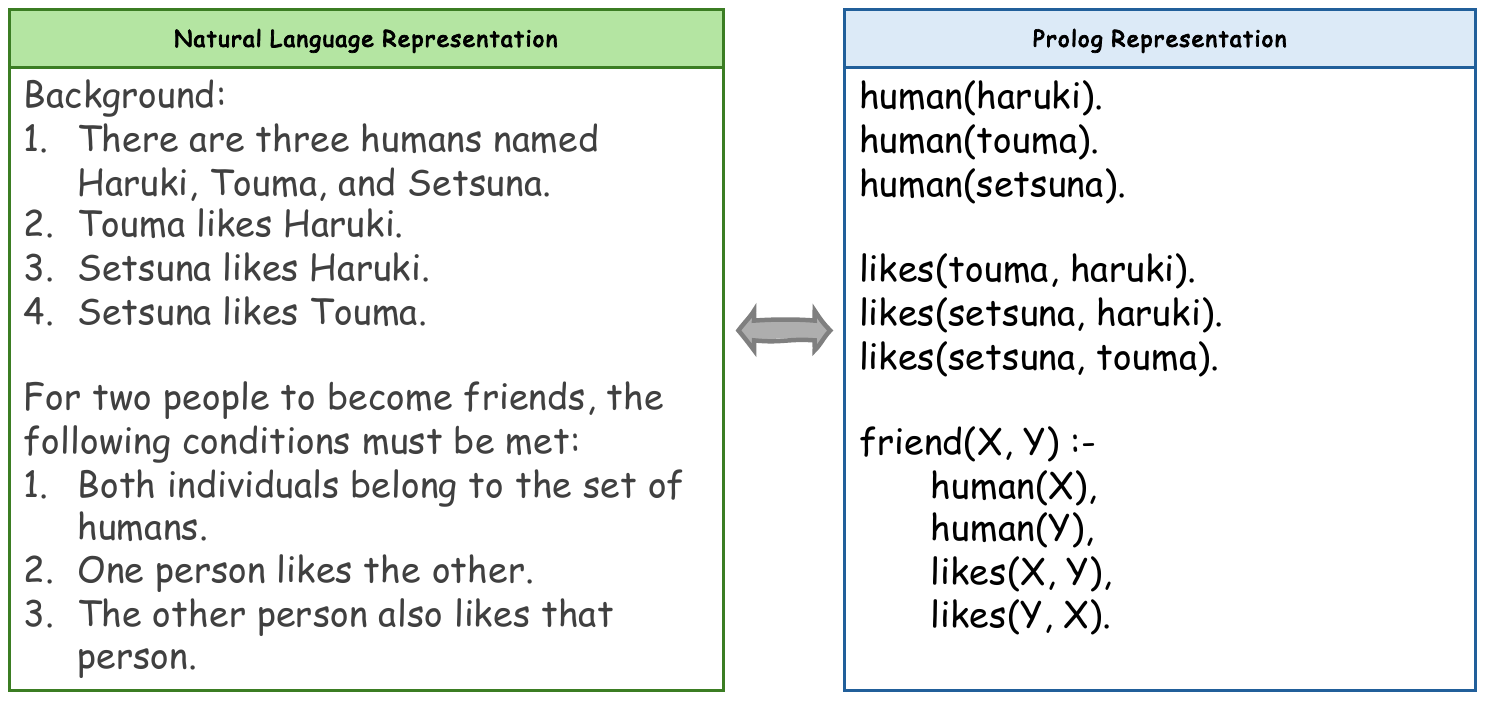}
    \caption{
    \textbf{Prolog Example}: Prolog representation divides logical problems into facts and rules, closely resembling natural language while preserving the logical structure of reasoning problems. More information about the features of Prolog and details of PDDL  is available in Appendix \ref{appendix:prolog_intro} and Appendix \ref{appendix:pddl_intro}.   }
    \label{fig:prolog_example}
\end{figure*}

To validate this hypothesis and leverage the isomorphic reasoning process to enhance reasoning capabilities, we introduce the \textbf{ProtoReasoning} framework, as illustrated in \cref{fig:overview}.
This framework leverages powerful \textbf{prototype systems} (abstract expressive representation together with a verification system) to formally represent reasoning patterns. Training models to solve problems encoded in abstract prototypes improves performance on a wide range of problems that share similar underlying reasoning structures.
We validate the effectiveness of ProtoReasoning in two core domains: logical reasoning and planning, employing Prolog (Programming in Logic)~\cite{covington1994prolog} as the prototype representation for logical reasoning and PDDL (Planning Domain Definition Language)~\cite{aeronautiques1998pddl} for planning tasks.
These languages serve as ideal prototypes due to three key characteristics: 
(1) \textbf{Declarative nature} --- Both focus on problem specification instead of procedural implementation, maintaining the reasoning structure found in natural language, as shown in \cref{fig:prolog_example}; 
(2) \textbf{Expressiveness} --- Prolog captures relational reasoning and constraint satisfaction through first-order predicate logic, while PDDL formalizes state transition systems for sequential planning; 
(3) \textbf{Verifiability} --- both possess mature interpreters (SWI-Prolog~\cite{wielemaker2012swi} and VAL~\cite{howey2004val}) enabling rigorous verification of reasoning chains.

Building on these foundational representations, our methodology establishes a comprehensive pipeline that includes dataset construction, verification, and model training.
For logical reasoning, we develop an automated workflow that converts natural language problems into Prolog representations with interpreter-verified solutions.
For planning domains, we introduce three novel task formulations (Plan Generation, Plan Completion, and Plan Reordering) with specialized verification systems ensuring correctness. 

The key advantage of our approach is leveraging the strengths of prototype systems, which incorporate comprehensive representation capabilities accompanied with built-in verification mechanisms. 
These properties enable us to automatically generate diverse problems within the prototype space while guaranteeing their correctness. 
Without the demand for problem-answer pairs, we only need executable prototype code representing the problem and can derive solutions via an interpreter. 
Thus, this approach eliminates annotation requirements, providing significant scaling potential.

We employ Supervised Fine-Tuning (SFT) on large language models to confirm the effectiveness of our methodology.
Compared to baseline, ProtoReasoning achieves a 4.7\% performance improvement on the Enigmata-Eval~\citep{chen2025enigmatascalinglogicalreasoning} logical reasoning benchmark, a 6.3\% improvement on planning tasks, 4.0\% improvement on MMLU~\citep{hendrycks2021mmlu} general reasoning benchmark and 1.0\% on AIME24~\citep{aime24} mathematical reasoning.
Significantly, our ablation experiments confirm that training in reasoning prototypes achieves transfer performance comparable to training on structurally similar problems in natural language, demonstrating that reasoning prototypes form the foundation for cross-domain reasoning generalization.

Our contributions are threefold: 
\begin{enumerate}
\item We introduce the concept of reasoning prototypes as a foundation for understanding cross-domain generalization phenomena in reasoning models.
\item We propose ProtoReasoning: an automated, scalable framework with integrated verification for enhancing reasoning and planning through systematic data construction.
\item We empirically demonstrate that training on abstract prototype representations significantly enhances models' ability to transfer reasoning capabilities across structurally similar problems.
\end{enumerate}
\section{The ProtoReasoning Framework}

\label{sec:approach}
\subsection{Overview}
This section introduces the ProtoReasoning framework, which utilizes representative prototype representations to enhance the general reasoning capabilities of LLMs.

The ProtoReasoning framework consists of two key modules:
\begin{itemize}
\item  \textbf{Prototype Constructor}: Responsible for transforming problems into corresponding prototype representations.
\item  \textbf{Verification System}: Responsible for evaluating the correctness of model outputs in the prototype representation space.
\end{itemize}

In the following sections, we will detail Prolog and PDDL as typical prototype representations for logical reasoning and planning tasks.

\subsection{Logic Prototype Representations}
Prolog (Programming in Logic)~\cite{covington1994prolog}, a declarative language founded on first-order predicate logic~\cite{enderton2001mathematical-logic-fol}, employs unification and backtracking mechanisms to represent and solve logical problems. Its declarative nature enables the expression of problems as logical constraints rather than procedural algorithms, creating a natural correspondence with human reasoning patterns. By training LLMs to solve Prolog-formulated problems, we systematically enhance their fundamental logical reasoning capabilities across diverse problem domains.

Next, we introduce our Prolog-based Logic Prototype Constructor and Verification System.

\subsubsection{Prolog-based Logic Prototype Constructor}
We designed a four-stage, model-driven pipeline to create diverse, verifiable Prolog reasoning problems:
\begin{itemize}
    \item [1)] \textbf{Data Initialization}. We collected comprehensive corpora of reasoning problems from web sources, encompassing both structured question-answer pairs and unstructured logical narratives.
    \item [2)] \textbf{Prototype Transformation}. Through prompt engineering with LLMs, we converted natural language logic problems into formal Prolog representations, while standardizing interpreter outputs as structured JSON format.
    \item [3)]  \textbf{Data Evolution}. We leveraged prompt engineering to systematically enhance problem complexity in a controllable manner while preserving JSON output constraints. 
    \item [4)] \textbf{Answer Derivation}. We employed the SWI-Prolog interpreter~\cite {wielemaker2008swi-prolog} to derive ground-truth answers. This interpreter-based approach eliminates reliance on pre-existing question-answer pairs and allows us to confidently evolve problem difficulty without introducing answer inaccuracies.
\end{itemize}

The resulting dataset consists of formalized problem-answer pairs:
\begin{align}
\mathcal{D}_\mathrm{Prolog} = \left\{ {\langle \mathcal{Q}_{\mathrm{Prolog}}, \mathcal{A} \rangle}_{i} \right\},
\end{align}

Where each instance $i$ contains a logical problem $\mathcal{Q}_{\mathrm{Prolog}}$ formulated in Prolog representation, paired with its corresponding verified answer $\mathcal{A}$ derived from the Prolog interpreter. We present the prompt templates used for prototype transformation and data evolution in the \cref{appendix:pe_template_prolog}.







\subsubsection{Prolog-based Verification System}
Building upon the prototype transformation described in the preceding section, we constrain all reference answers $\mathcal{A}$ to be structured JSON dictionaries.
We then developed a training prompt template that instructs models to generate predictions $\hat{\mathcal{A}}$ in the same JSON format.
This standardization facilitates rigorous evaluation between the outputs of the Prolog interpreter and model predictions, ensuring both are represented in a consistent format.

Our training prompt template is shown below, where "\texttt{[program]}" and "\texttt{[query]}" are replaced with the corresponding values for each sample:
\begin{tcolorbox}[
    title=Prolog Execution Prompt Template for Training,
    colback=lightblue!10,
    colframe=seedblue!70,
    coltitle=white,
    boxrule=1pt,
    arc=2mm,
    breakable,
    left=5pt,
    right=5pt,
    top=5pt,
    bottom=5pt
]
\small
\begin{ttfamily}
\#\# Prolog Execution Parser

As a Prolog code execution expert, your task is to simulate the execution process of SWI-Prolog version 8.2.4 in a Linux x86\_64 environment.

\#\# Workflow

You will receive a Prolog program and query, then return the execution results, simulating the following command line operation: swipl -q -f prolog\_program.pl -t solve\_json,halt.

\#\# Technical Specifications

Results must 100\% accurately reflect the actual output of SWI-Prolog.

\#\# Input

[Program]

[Query]

\#\# Output Rules

Provide only the execution results in JSON format, wrapped with JSON and markers.
\end{ttfamily}
\end{tcolorbox}
\subsection{Planning Prototype Representations}
PDDL (Planning Domain Definition Language)~\cite{aeronautiques1998pddl} is the standard representation for automated planning problems, modeling state transition systems through three essential components: state representations, actions with preconditions and effects, and state transitions. This representation naturally aligns with human planning cognition, particularly in reasoning about action requirements and consequences.

Next, we will explain how to construct prototype learning for planning capabilities based on PDDL.

\subsubsection{PDDL-based Planning Prototype Constructor}
For planning tasks, we employed PDDL-Generator \cite{seipp-et-al-zenodo2022-pddl-generator} and FastDownward\cite{Helmert_2006} to create problems and derive optimal solutions across diverse domains. Based on International Planning Competition (IPC) \cite{IPC} benchmarks, we generated varying complexity problems in classical domains like BlocksWorld and Logistics. For each domain, we constructed three distinct task types:
\begin{itemize}
    \item [1)] \textbf{Plan Generation}: Given a domain definition and problem description, the model must generate a complete sequence of actions from initial to goal state, testing end-to-end planning ability.

    \item [2)] \textbf{Plan Completion}: Provided with a partial action sequence, the model must fill in missing steps, requiring an understanding of intermediate states and bidirectional reasoning.

    \item [3)] \textbf{Plan Reordering}: Given unordered action steps, the model must determine a valid execution sequence, testing comprehension of action dependencies and precondition relationships.
\end{itemize}
The resulting dataset is structured as follows:
\begin{align}
\mathcal{D}_\mathrm{PDDL} = \left\{ {\langle \mathcal{Q}_{\mathrm{PDDL}}, \mathcal{P}_\mathrm{ref} \rangle}_{i} \right\}, 
\end{align}

where each instance $i$ contains a planning problem formulated in PDDL representation $\mathcal{Q}_{\mathrm{PDDL}}$, paired with its corresponding task-specific reference $\mathcal{P}_\mathrm{ref}$. These references vary according to task type, with differences explained in detail in \cref{sec:pddl_verification_system}.

\subsubsection{PDDL-based Verification System}\label{sec:pddl_verification_system}
Unlike Prolog verification which relies on exact JSON dictionary comparison, PDDL tasks require specialized validation as multiple valid plans can exist for the same problem. Leveraging VAL (PDDL Plan Validator), we implemented custom verification procedures for each task type:
\begin{itemize}
    \item [1)] \textbf{Plan Generation}: For standard generation tasks, we accept any plan that passes VAL verification as correct, with no reference values required. For optimization tasks where plan cost matters (minimum or maximum cost plans), we use FastDownward solver~\cite{Helmert_2006} to establish reference optimal values. In these cases, a generated plan is considered correct only if it both passes VAL verification and achieves the required optimality criteria.

    \item [2)] \textbf{Plan Completion}: Here, $\mathcal{P}_\mathrm{ref}$ contains the given partial plan. The model's output must pass VAL verification while incorporating all actions from the partial plan in their exact positions.

    \item [3)] \textbf{Plan Reordering}: In this case, $\mathcal{P}_\mathrm{ref}$ contains unordered action steps. A correct solution must pass VAL verification and contain the same set of actions as specified in $\mathcal{P}_\mathrm{ref}$ --- no missing actions and no additional ones.
\end{itemize}

Similar to our Prolog approach, we employed three training prompt templates to guide model generation in verifiable formats. Details are available in the \cref{appendix:pe_for_pddl}.

\subsection{Model Training Recipe}
In this section, we introduce our training approach that utilizes supervised fine-tuning to validate effectiveness. Our methodology employs a three-phase process:
\begin{itemize}
    \item [1)] \textbf{Teacher Model Distillation}: We employed Deepseek-R1 \cite{guo2025deepseek}, a high-performance language model, to generate explicit reasoning chains for our initial datasets. This transformation enriched our data to include step-by-step reasoning paths, creating augmented datasets $\mathcal{D}_\mathrm{Prolog}^{\mathrm{Aug}} = \left\{ {\langle \mathcal{Q}_{\mathrm{Prolog}}, \mathcal{R}_\mathrm{CoT}, \mathcal{A} \rangle}_{i} \right\}$ for Prolog tasks and $\mathcal{D}_\mathrm{PDDL}^{\mathrm{Aug}} = \left\{ {\langle \mathcal{Q}_{\mathrm{PDDL}}, \mathcal{R}_{\mathrm{CoT}}, \mathcal{P}_\mathrm{ref} \rangle}_{i} \right\}$ for PDDL tasks. We subsequently fine-tuned our base model using these datasets.

    \item [2)] \textbf{Difficulty Stratification}: Using the model trained in the previous phase, we implement a rejection sampling methodology to classify instances by difficulty. We evaluate each problem 10 times and categorize it based on pass rate:
    \begin{align}
        \begin{cases}
            \text{Challenging} & \text{if } 0 < \text{Pass Rate} \leq 0.3 \\
            \text{Intermediate} & \text{if } 0.4 \leq \text{Pass Rate} \leq 0.6 \\
            \text{Elementary} & \text{if } 0.7 \leq \text{Pass Rate} < 1 
        \end{cases}
    \end{align}
    We exclude both perfectly solved instances (pass rate = 1.0) and completely failed ones (pass rate = 0), then train an enhanced model on this stratified dataset.

    \item [3)] \textbf{Quality Filtration}: With our refined model from the second phase, we conduct a final round of rejection sampling to create our definitive training dataset.
\end{itemize}
\section{Experiments}

\subsection{Experimental Setup}

\begin{wraptable}{r}{0.4\linewidth}
\vspace{-1.0em}
\small
\centering
\caption{Hyperparameters for Model Training}
\label{tab:parameters}
    \begin{tabular}{@{}ll@{}}
    \toprule
    \textbf{Parameter} & \textbf{Value} \\
    \midrule
    \multicolumn{2}{@{}l}{\textit{Model Architecture}} \\
    \midrule
    Total parameters & 150B \\
    Activated parameters & 15B \\
    \midrule
    \multicolumn{2}{@{}l}{\textit{Training Configuration}} \\
    \midrule
    Learning rate & 2e-5 \\
    Batch size & 6 \\
    Sequence packing & Yes \\
    Optimizer & AdamW \\
    Weight decay & 0.1 \\
    Epoch & 2\\
    Warmup step rate & 0.01 \\
    Learning rate schedule & Cosine decay \\
    Maximum sequence length & 32768 \\
    \bottomrule
    \end{tabular}
\vspace{-1.0em}
\end{wraptable}

\textbf{Dataset}.
Our baseline dataset comprises 100K diverse samples from the Seed Project, covering various tasks including mathematical reasoning, code generation, and creative writing. Applying our methodology, we initially collected 43,021 raw Prolog examples and 34,123 raw PDDL instances. After processing these through the three-phase pipeline described in our Training Recipe, we refined our collection to 4,196 high-quality Prolog problems and 2,424 PDDL tasks for the final training dataset.

\textbf{Evaluation}.
We evaluated logical reasoning capabilities using the Enigmata-Eval benchmark \citep{chen2025enigmatascalinglogicalreasoning}. To focus specifically on logical reasoning improvements rather than instruction-following abilities, we excluded four datasets (Campsite, Car Painting, Star Battle, and Sum Skyscraper) where baseline models struggled primarily with following instructions rather than the reasoning tasks themselves. 
For planning assessment, we employed a dual approach: direct evaluation using an internal planning test set from the Seed Project, and indirect function calling evaluation through the Nexus-Hard benchmark (another internal evaluation set representing a more challenging subset of Nexus~\citep{nexusraven2}). 
This complementary evaluation strategy leverages the cognitive overlap between planning and function calling --- both requiring sequential action formulation and parameter management --- to provide insights into the cross-domain transfer of planning abilities. 
To assess out-of-domain generalization, we evaluate ProtoReasoner on MMLU~\citep{hendrycks2021mmlu} (general reasoning) and AIME24~\citep{aime24} (mathematical reasoning), extending beyond its native logic and planning domains.
To ensure result stability, we evaluated each sample 3 times and used the average score as the final result, except AIME2024, where each sample was evaluated 10 times.
All benchmarks are 0-shot evaluations.

\textbf{Hyperparameters}.
We adopt a Mixture-of-Experts (MoE) architecture that activates 15B parameters from a total parameter count of 150B as our training model. 
The training employed a learning rate of 2e-5 and a batch size of 6 with sequence packing enabled.
The complete experimental hyperparameters are presented in~\cref{tab:parameters}.


\begin{table}[!tb]
\centering
\caption{
Performance of baseline and ProtoReasoning on different benchmarks. 
Results demonstrate that our ProtoReasoning framework improves both logical reasoning and planning ability, showing a generalization from prototype representation to natural language representation.
Scores represent the average performance across all samples in each test set. 
}
\label{tab:results}
\begin{tabular}{@{}l|lllll@{}}
\toprule
\textbf{Method} & \textbf{Enigmata-Eval} & \textbf{Nexus-Hard} & \textbf{Task Planning} & \textbf{AIME2024} & \textbf{MMLU} \\
\midrule
\midrule
\textbf{Baseline}        & 37.3 & 53.1 & 46.7 & 72.0 & 82.7 \\
\textbf{ProtoReasoning}  & $42.0_{\textcolor{ForestGreen}{\uparrow 4.7\%}}$ & $59.5_{\textcolor{ForestGreen}{\uparrow 6.4\%}}$ & $53.0_{\textcolor{ForestGreen}{\uparrow 6.3\%}}$ & $73.0_{\textcolor{ForestGreen}{\uparrow 1.0\%}}$ & $86.7_{\textcolor{ForestGreen}{\uparrow 4.0\%}}$\\
\bottomrule
\end{tabular}
\end{table}

\subsection{Experimental Results}
\begin{table}[!tb]
    \centering
    \caption{
    Performance comparison between the baseline method and ProtoReasoning across Enigmata-Eval categories. ProtoReasoning achieves improvements in all tasks, with the largest gain in Crypto ($\textcolor{ForestGreen}{+11.0\%}$) and the smallest in Sequential ($\textcolor{ForestGreen}{+0.3\%}$).
    }
    \begin{tabular}{l|llllllll}
        \toprule
        \textbf{Method} & \textbf{Arith.} & \textbf{Crypto} & \textbf{Graph} & \textbf{Search} & \textbf{Seq.} & \textbf{Grid} & \textbf{Logic} \\
        \midrule \midrule
        \textbf{Baseline} & 61.0 & 28.8 & 43.4 & 26.1 & 16.0 & 43.8 & 65.3 \\
        \textbf{ProtoReasoning} & $64.7_{\textcolor{ForestGreen}{\uparrow 3.7\%}}$ & $39.8_{\textcolor{ForestGreen}{\uparrow 11.0\%}}$ & $52.4_{\textcolor{ForestGreen}{\uparrow 9.0\%}}$ & $29.3_{\textcolor{ForestGreen}{\uparrow 3.2\%}}$ & $16.3_{\textcolor{ForestGreen}{\uparrow 0.3\%}}$ & $48.2_{\textcolor{ForestGreen}{\uparrow 4.4\%}}$ & $68.6_{\textcolor{ForestGreen}{\uparrow 3.3\%}}$ \\
        \bottomrule
    \end{tabular}
    \label{tab:model_performance}
\end{table}

As demonstrated in \cref{tab:results}, incorporating synthetic Prolog and PDDL data led to substantial improvements across all reasoning benchmarks. On the Enigmata-Eval benchmark, our approach increased logical reasoning performance from 37.3\% to 42.0\%, representing a 4.7\% improvement. 
This significant gain suggests that training with logic prototypes enhances the model's ability to recognize and apply fundamental reasoning patterns. Detailed performance breakdowns by category are available in \cref{tab:model_performance}.

Similarly, planning capabilities showed marked enhancement, with Nexus-Hard scores rising from 53.1\% to 59.5\% and dedicated planning task performance increasing from 46.7\% to 53.0\%. 
The consistent improvements across both direct planning tasks and function calling evaluations indicate that the skills acquired through prototype-based training are transferred effectively across different manifestations of planning problems. 


Notably, ProtoReasoning demonstrates strong generalization beyond its core training domains: compared to standard LLM training, it elevates performance on the general knowledge benchmark MMLU~\citep{hendrycks2021mmlu} from 82.7\% to 86.7\% (+4.0\%), and on the mathematical reasoning benchmark AIME2024~\citep{aime24} from 72.0\% to 73.0\% (+1.0\%). These gains confirm ProtoReasoning's effectiveness not only in logical reasoning and planning domains but also in general knowledge reasoning and mathematical reasoning.

\subsection{Ablation Study}
This section presents a detailed analysis of our proposed reasoning prototypes, examining their effectiveness at a fundamental level.

\textbf{Ablation Setup}.
For our ablation study, the Enigmata-Eval benchmark~\citep{chen2025enigmatascalinglogicalreasoning} was partitioned into distinct training and development sets. 
We processed the training subset using two different methods:
(1) conversion into Prolog representations with interpreter-verified solutions, and
(2) preparation of the original natural language problems utilizing multi-stage rejection sampling validated by Enigmata-Eval's native verifier.
This procedure generated a matched training corpus of 453 samples, representing the intersection of both methods where the problems are encoded in both Prolog and natural language formats. This controlled experimental design isolates the impact of the representational format while keeping the problem content invariant between experimental configurations.

We design the following three experimental configurations:
\begin{enumerate}
\item \textbf{Baseline (Method 1)}: Model trained exclusively on our standard dataset (absent Enigmata-related problems).
\item \textbf{Baseline w/ Prolog Representation (Method 2)}: Model trained on standard dataset augmented with formalized Prolog representations of Enigmata reasoning problems, enabling evaluation of reasoning prototype effectiveness in enhancing reasoning capabilities.
\item \textbf{Baseline w/ Natural Language (Method 3)}: Model trained on the standard dataset augmented with natural language versions of the same reasoning problems used in the prototype approach, enabling direct comparison between prototypes and natural language representations.
\end{enumerate}

\begin{table}[!tb]
\centering
\caption{
Performance comparison between prototype and natural language representations on the Enigmata-Eval benchmark. The Prolog prototype achieves performance comparable to natural language, demonstrating its generalizable reasoning ability.
}
\label{tab:prototype_analysis}
\begin{tabular}{@{}l|ll@{}}
\toprule
\textbf{Method} & \textbf{Trans. Set} & \textbf{Dev. Set} \\
\midrule
\midrule
Baseline & 35.2 & 38.5 \\
  + Prolog & $54.2_{\textcolor{ForestGreen}{\uparrow 19.0\%}}$ & $44.1_{\textcolor{ForestGreen}{\uparrow 5.6\%}}$ \\
 + NL & $58.1_{\textcolor{ForestGreen}{\space \uparrow 22.9\%}}$ & $45.0_{\textcolor{ForestGreen}{\uparrow 6.5\%}}$ \\
 + Prolog (without CoT) & $41.9_{\textcolor{ForestGreen}{\uparrow 6.7\%}}$ & $39.6_{\textcolor{ForestGreen}{\space \uparrow 1.1\%}}$ \\
\bottomrule
\end{tabular}
\end{table}

\begin{table}[!tb]
\centering
\caption{
Prolog prototype achieves performance comparable to natural language (NL) representation across most categories in Enigmata-Eval.
However, it underperforms the baseline on the logic transfer subset, mainly attributed to limited samples.
For those categories contain sufficient samples, the performance of Prolog prototype match and can even surpass natural language representation, demonstrating stronger generalization capability.
}
\begin{tabularx}{0.90\columnwidth}{@{} 
    >{\centering\arraybackslash}X
    |
    >{\centering\arraybackslash}X
    ||
    *{3}{>{\centering\arraybackslash}X}
    ||
    >{\centering\arraybackslash}X
  @{}}
\toprule
\multicolumn{2}{c}{\textbf{Dataset}}  & Baseline & +NL & +Prolog & \textbf{Size} \\
\midrule
\midrule
\multirow{2}{*}{\textbf{Arith.}} & \textbf{Trans. Set} &  $48.5$ & $60.6_{\textcolor{ForestGreen}{\uparrow 12.1 \%}}$ & $56.1_{\textcolor{ForestGreen}{\uparrow 7.6 \%}}$ & {22} \\ 
  & \textbf{Dev. Set} & $62.0$ & $67.1_{\textcolor{ForestGreen}{\uparrow 5.1 \%}}$ & $69.4_{\textcolor{ForestGreen}{\uparrow 7.4 \%}}$ & {278} \\ \midrule
  
\multirow{2}{*}{\textbf{Crypto}} & \textbf{Trans. Set}  & $22.8$ & $48.9_{\textcolor{ForestGreen}{\uparrow 26.1 \%}}$ & $51.1_{\textcolor{ForestGreen}{\uparrow 28.3 \%}}$ & {120} \\
  & \textbf{Dev. Set}  & $32.8$ & $54.4_{\textcolor{ForestGreen}{\uparrow 21.6 \%}}$ & $44.4_{\textcolor{ForestGreen}{\uparrow 11.6 \%}}$ & {180}\\ \midrule
  
\multirow{2}{*}{\textbf{Graph}} & \textbf{Trans. Set} & $61.5$ & $80.5_{\textcolor{ForestGreen}{\uparrow 19.0 \%}}$ & $76.9_{\textcolor{ForestGreen}{\uparrow 15.4 \%}}$ & {65}\\
  & \textbf{Dev. Set}  & $39.9$ & $44.9_{\textcolor{ForestGreen}{\uparrow 5.0 \%}}$ & $45.8_{\textcolor{ForestGreen}{\uparrow 5.9 \%}}$ & {635} \\ \midrule

\multirow{2}{*}{\textbf{Search}} & \textbf{Trans. Set} & $41.1$ & $68.2_{\textcolor{ForestGreen}{\uparrow 27.1 \%}}$ & $58.1_{\textcolor{ForestGreen}{\uparrow17.0 \%}}$ & {43} \\ 
    & \textbf{Dev. Set} & $25.7$ & $35.7_{\textcolor{ForestGreen}{\uparrow 10.0 \%}}$ & $34.4_{\textcolor{ForestGreen}{\uparrow 8.7 \%}}$ & {757} \\ \midrule
    
\multirow{2}{*}{\textbf{Seq.}} & \textbf{Trans. Set} & $8.3$ & $25.0_{\textcolor{ForestGreen}{\uparrow 16.7 \%}}$ & $25.0_{\textcolor{ForestGreen}{\uparrow 16.7 \%}}$ & {4} \\ 
    & \textbf{Dev. Set} & $17.9$ & $18.3_{\textcolor{ForestGreen}{\uparrow 0.4 \%}}$ & $18.4_{\textcolor{ForestGreen}{\uparrow 0.5 \%}}$ & {804} \\ \midrule
    
\multirow{2}{*}{\textbf{Grid}} & \textbf{Trans. Set} & $16.7$ & $41.7_{\textcolor{ForestGreen}{\uparrow 25.0 \%}}$ & $38.6_{\textcolor{ForestGreen}{\uparrow 21.9 \%}}$ & {44}\\ 
    & \textbf{Dev. Set}  & $48.8$ & $55.6_{\textcolor{ForestGreen}{\uparrow 6.8 \%}}$ & $54.9_{\textcolor{ForestGreen}{\uparrow 6.1 \%}}$ & {756} \\ \midrule

\multirow{2}{*}{\textbf{Logic}} & \textbf{Trans. Set} & $39.6$ & $58.3_{\textcolor{ForestGreen}{\uparrow 18.7 \%}}$ & ${38.5_{\textcolor{red}{\downarrow 1.1 \%}}}$ & {32} \\ 
    & \textbf{Dev. Set} & $67.3$ & $75.7_{\textcolor{ForestGreen}{\uparrow 8.4 \%}}$ & $72.2_{\textcolor{ForestGreen}{\uparrow 4.9 \%}}$ & {418}\\
\bottomrule
\end{tabularx}
\label{tab:ablation_detail_category}
\end{table}

\textbf{Ablation Evaluation}. We evaluated model performance using two complementary test sets to assess different aspects of reasoning transfer. Similar to previous experiment, we sample each problem three times and used the average score for reliable assessment:

\begin{enumerate}
\item \textbf{Prototype Transfer Set}: This comprises the original natural language versions of problems in Enigmata-Eval used to create our Prolog training examples. Performance in this set measures how effectively reasoning capabilities acquired through logic prototype training transfer to solving natural language problems with similar logical structures.

\item \textbf{Development Set}: This consists of the remaining portion of Enigmata-Eval that was not used in creating the Prototype Transfer Set. Performance in this set evaluates broader generalization capabilities, assessing whether models can apply learned reasoning patterns to problems outside of the training distribution.
\end{enumerate}

\textbf{Results and Analysis}. 
\cref{tab:prototype_analysis} demonstrates that Prolog representations yield significant and consistent performance gains across both transfer and development sets. 
The performance gain of the baseline w/ Prolog representation (Method 2) over the baseline (Method 1) validates that training in the prototype representation effectively generalizes to natural language problems. 
Moreover, the comparable performance between method 2 (w/ Prolog representation) and method 3 (w/ natural language) confirms that prototype training achieves effect similar to direct natural language training, consolidating the generalization capability of the prototype.
Additionally, our Prolog training experiment without CoT reasoning showed dramatically reduced performance, confirming that effective generalization through prototypes requires explicit reasoning processes, as they benefit from homogenized reasoning paths. 

\cref{tab:ablation_detail_category} details category-wise performance for both transfer and development sets. 
Insufficient samples in the transfer set of logic category resulted in unstable evaluation, yielding performance worse than baseline. 
However, in most cases with sufficient samples, training in the Prolog prototype achieved comparable or even superior performance to training in natural language representation.
Notably, our analysis uses sample-averaged results rather than category-specific metrics due to our limited training dataset. 
Since our transfer set consists of the natural language versions of the same problems used in the Prolog training set, sample-averaged metrics are more reliable than category-specific analysis.

\section{Related Work}

\textbf{Long CoT and Reasoning Model}.
Recent advances in LLM reasoning, such as OpenAI-o1~\citep{openai-o1}, DeepSeek-R1~\citep{guo2025deepseek}, Seed-Thinking-v1.5~\citep{seed2025seed-thinking} and Kimi-k1.5~\citep{team2025kimi}, have changed the focus from Chain-of-Thought (CoT)~\citep{wei2022chain-of-thought} and supervised fine-tuning (SFT) to reinforcement learning (RL).
Deepseek-R1~\citep{guo2025deepseek} leveraged mathematical problems and code executions to increase Long CoT reasoning capabilities. Seed-Thinking-v1.5~\citep{seed2025seed-thinking} employed various task collections that span mathematics and logic puzzles. Logic-RL~\citep{xie2025logic-rl} utilized Knights and Knaves puzzles~\citep{xie24memorization-kk-puzzle} to demonstrate generalization to challenging mathematical benchmarks. In this stage, RL algorithms, e.g., PPO~\citep{schulman2017ppo}, GRPO~\citep{shao2024deepseekmath-grpo}, DAPO~\citep{yu2025dapo}, are adopted to guide the LLM exploring reasoning paths and to stimulate the long CoT reasoning ability, relying on verifiable rewards, such as accuracy based on ground-truth answers.
These approaches collectively demonstrate the effectiveness of reinforcement learning with verifiable reward (RLVR)~\citep{kumar2025llm-posttrain-survey} in developing sophisticated reasoning abilities, including strategies such as recognizing correcting mistakes, breaking down difficult steps and iterating on alternative approaches, thus showcasing the powerful generalization capacity of long Chain-of-Thought reasoning~\citep{chen2025longcot-survey}.
Compared to previous work, our work introduces the concept of reasoning prototypes, aimed at understanding the underlying generalization mechanisms emerging from long chain-of-thought, and provides a more fundamental framework for cross-domain reasoning transfer. 

\textbf{Symbolic Reasoning in LLMs}. Large language models conduct reasoning not only in natural language space~\citep{wei2022chain-of-thought} but also in a neuro-symbolic manner~\citep{gao2023pal, ye2023satlm, pan2023logic-lm}. The intermediate reasoning steps can manifest themselves as code~\citep{gao2023pal}, domain-specific languages~\citep{ye2023satlm, zhou2024dtv}, or mixtures of different symbolic representations~\citep{zheng2025learning-mot, han2025hybridmind}. 
For Prolog programming language, some previous work~\citep{covington1994prolog, yang2024arithmetic-prolog, borazjanizadeh2024reliable-prolog} leveraged it as an intermediate representation to improve the reasoning ability of LLMs. 
Rather than emphasizing the specific manifestation of the Chain-of-Thought process, we investigate how prototypes can stimulate effective reasoning by exploiting the inherent thinking structures between tasks.
\section{Conclusion and Future Work}
This paper introduces ProtoReasoning, a framework that validates the hypothesis that abstract reasoning prototypes serve as the foundation for cross-domain generalization in large language models. By training on prototype representations (Prolog for logical reasoning, PDDL for planning), we achieve significant improvements on both logical reasoning and planning tasks, with ablation studies confirming effective transfer to structurally similar problems. Additionally, we believe this framework is generalizable for other LLM capabilities. 
However, our theoretical understanding remains insufficient --- the precise definition of "reasoning prototypes" lacks formal rigor, and the underlying mechanisms driving cross-domain transfer require deeper investigation. 
In the future, we should develop more rigorous mathematical frameworks to characterize these prototypes and provide stronger theoretical foundations for our empirical findings.
We will also open-source the curated prototype datasets (Prolog and PDDL) used in this study to accelerate research progress in the community. 
Further, we will reproduce our result in open-sourced large language models~\citep{yang2025qwen3, qwen2025qwen25technicalreport} to ensure a broader validation of our hypothesis.


\bibliographystyle{plainnat}
\bibliography{main}

\clearpage

\beginappendix

\section{Prompt Template for Training Planning Tasks}
\label{appendix:pe_for_pddl}

\begin{tcolorbox}[
    title=PDDL Plan Generation Prompt Template,
    colback=lightblue!10,
    colframe=seedblue!70,
    coltitle=white,
    boxrule=1pt,
    arc=2mm,
    breakable,
    left=5pt,
    right=5pt,
    top=5pt,
    bottom=5pt
]
\small
Now, as an authoritative expert in the field of PDDL planning, you will face an important challenge. I will provide you with a PDDL domain file that defines the basic framework and available operations for problem-solving, and a problem description file that clarifies the initial state and the target state. Your core mission is to use your professional knowledge, analyze these two files, and design a PDDL planning solution.

\textbf{\#\#\#Input}

\textbf{[PDDL Domain File]}\\
\texttt{```pddl}\\
\texttt{\{Specific content of the PDDL domain definition\}}\\
\texttt{```}

\textbf{[PDDL Problem File]}\\
\texttt{```pddl}\\
\texttt{\{Detailed description content of the PDDL problem\}}\\
\texttt{```}

\textbf{\#\#\#Output}

Please output the finally generated planning solution in the strict PDDL format and enclose it with the tags \texttt{```pddl} and \texttt{```} for clearly and accurately presenting the complete plan. An example is as follows:\\
\texttt{```pddl}\\
\texttt{\{valid plan\}}\\
\texttt{```}
\end{tcolorbox}
\begin{tcolorbox}[
    title=PDDL Planning Completion Prompt Template,
    colback=lightblue!10,
    colframe=seedblue!70,
    coltitle=white,
    boxrule=1pt,
    arc=2mm,
    breakable,
    left=5pt,
    right=5pt,
    top=5pt,
    bottom=5pt
]
\small
Please act as a PDDL planning expert. I will provide a Domain, a Problem, and a Partial Plan obtained by randomly deleting parts from a complete and valid Plan. Your task is to restore the complete Plan based on this information.

\textbf{\#\#\# Input}

\textbf{[Domain]}\\
\texttt{```pddl}\\
\texttt{\{pddl\_domain\}}\\
\texttt{```}

\textbf{[Problem]}\\
\texttt{```pddl}\\
\texttt{\{pddl\_problem\}}\\
\texttt{```}

\textbf{[Partial Plan]}\\
\texttt{```pddl}\\
\texttt{\{partial\_plan\}}\\
\texttt{```}

\textbf{\#\#\# Output}

Please present the plan in PDDL format and enclose it with the tags \texttt{```pddl} and \texttt{```}, for example:\\
\texttt{```pddl}\\
\texttt{\{valid plan\}}\\
\texttt{```}
\end{tcolorbox}

\begin{tcolorbox}[
    title=PDDL Plan Reordering Prompt Template,
    colback=lightblue!10,
    colframe=seedblue!70,
    coltitle=white,
    boxrule=1pt,
    arc=2mm,
    breakable,
    left=5pt,
    right=5pt,
    top=5pt,
    bottom=5pt
]
\small
From now on, assume the role of a senior expert in PDDL planning. I will provide you with three key pieces of information: a PDDL domain description, a PDDL problem definition, and a valid PDDL plan with its execution steps out of order. Your core task is to, based on the rules related to the domain and the problem, sort out the disordered planning steps and rearrange them into a logically coherent sequence that meets the execution requirements.

\textbf{\#\#\#Input}

\textbf{[Domain Description]}\\
\texttt{```pddl}\\
\texttt{\{pddl\_domain\}}\\
\texttt{```}

\textbf{[Problem Definition]}\\
\texttt{```pddl}\\
\texttt{\{pddl\_problem\}}\\
\texttt{```}

\textbf{[Out-of-Order Plan]}\\
\texttt{```pddl}\\
\texttt{\{output\_of\_order\_plan\}}\\
\texttt{```}

\textbf{\#\#\#Output}

Please output the complete plan with the adjusted sequence in the standard PDDL format. Be sure to enclose it with the tags \texttt{```pddl} and \texttt{```}. The example is presented as follows:\\
\texttt{```pddl}\\
\texttt{\{valid plan\}}\\
\texttt{```}
\end{tcolorbox}

\section{Prompt Engineering Template for Prolog}
\label{appendix:pe_template_prolog}
To systematically convert natural language problems into Prolog representations, we designed a structured prompt template that guides the transformation process:
\begin{tcolorbox}[
    title=Prolog Transformation Prompt Template,
    colback=lightblue!10,
    colframe=seedblue!70,
    coltitle=white,
    boxrule=1pt,
    arc=2mm,
    breakable,
    left=5pt,
    right=5pt,
    top=5pt,
    bottom=5pt
]
\small
\texttt{\#\# [Role]\\
You are an exceptional logical reasoning expert, skilled at deconstructing complex logical problems described in natural language and transforming them into precise formal expressions. As a Prolog language specialist, you can elegantly convert various logical problems into executable Prolog code, while ensuring complete logical consistency between the formalized expression and the original natural language description.\\
\#\# [Constraints]\\
1. Logic transformation must be precise and error-free, with transparent and complete process:\\
\hspace{1em}* Strictly prohibit presetting any conclusions in the code; answers must be derived through Prolog execution\\
\hspace{1em}* Programs avoid any logical jumps\\
\hspace{1em}* Information requiring calculation in the original text must show the complete calculation process in the code\\
2. Standardized queries:\\
\hspace{1em}* Use only solve\_json. as the query entry point\\
\hspace{1em}* Results must output key logical outcomes in a concise and clear JSON format\\
\hspace{1em}* Must import and correctly use :- use\_module(library(http/json)) module for JSON conversion\\
\hspace{1em}* You should ensure that if there is no valid answer, the JSON output is \{``result'': ``No valid solution found''\}.\\
\hspace{1em}* The final JSON string should be exclusively output to the standard output (stdout).(Like ``json\_write(current\_output, JsonTerm, [width(0)]).'')\\
3. Code quality requirements:\\
\hspace{1em}* Ensure code can be correctly executed by SWI-Prolog\\
\hspace{1em}* Clearly divide into Program and Query sections\\
\hspace{1em}* Code logic must be complete and correct, with no execution errors\\
4. Output specifications:\\
\hspace{1em}* Generated JSON must be compatible with Python3 json.loads parsing\\
\hspace{1em}* Prohibit outputting Prolog internal variables or intermediate results\\
\hspace{1em}* Appropriately convert data structures that are difficult to represent in JSON\\
\hspace{1em}* You should ensure that if there is no valid answer, the JSON output is \{``result'': ``No valid solution found''\}.\\
5. Increasing problem complexity:\\
\hspace{1em}* For simple multiple-choice questions, true/false questions, or other easily guessable problems, increase their complexity\\
\hspace{1em}* Ensure the result space is sufficiently large and not easily randomly guessed\\
6. Determinism:\\
\hspace{1em}* Code execution results must be deterministic; random logic is prohibited\\
\#\# [Input]\\
\{prompt\}\\
\#\# [Output]\\
Because Query can only be solve\_json. , you only need to output Program:}
\end{tcolorbox}

\begin{tcolorbox}[
    title=Prolog Problem Generalization Prompt Template,
    colback=lightblue!10,
    colframe=seedblue!70,
    coltitle=white,
    boxrule=1pt,
    arc=2mm,
    breakable,
    left=5pt,
    right=5pt,
    top=5pt,
    bottom=5pt
]
\small
\texttt{\#\# [Instructions]\\
You will continue the conversation above and complete the following functionality (please note that your generated code still needs to meet the format requirements mentioned above):\\
\#\# [I. Background]\\
* You need to generalize a logical or algorithmic problem, creating a more complex and challenging variant, and implement it in Prolog.\\
* Generalization means preserving the core idea of the original problem while making it more universal or complex by changing parameters, adding constraints, or modifying objectives.\\
\#\# [II. Role]\\
* You are a computer scientist specializing in logic programming, particularly skilled in the Prolog language and problem formalization.\\
* Your task is to creatively generalize problems while ensuring correctness and executability of their implementation.\\
\#\# [III. Generalization Requirements]\\
* The generalization should preserve the essence of the original problem while increasing its complexity or universality.\\
* The generalized problem should have a sufficiently large solution space, avoiding:\\
\hspace{1em}* Multiple-choice, true/false, or other easily guessable formats.\\
\hspace{1em}* Problems with extremely few solutions.\\
\hspace{1em}* Problems with overly obvious solution patterns.\\
* Generalization directions may include but are not limited to:\\
\hspace{1em}* Increasing problem scale (e.g., from 8-queens to n-queens).\\
\hspace{1em}* Modifying constraints (e.g., adding mandatory waypoints, resource limitations).\\
\hspace{1em}* Changing optimization goals (e.g., from shortest path to specific-length path).\\
\hspace{1em}* Adding extra dimensions (e.g., adding a time dimension to a 2D problem).\\
\#\# [IV. Examples]\\
* \textbf{Original problem}: Select numbers from a set of 10 to sum to 30.\\
\hspace{1em}* \textbf{Generalized}: Select combinations of items from a given list to have a total value within a specific range while not exceeding a backpack weight capacity.\\
* \textbf{Original problem}: 8-queens problem.\\
\hspace{1em}* \textbf{Generalized}: N-queens problem with certain squares pre-forbidden.\\
* \textbf{Original problem}: Shortest path between cities.\\
\hspace{1em}* \textbf{Generalized}: Delivery route optimization problem with time window constraints.\\
\#\# [V. Evaluation Criteria]\\
* \textbf{Creativity}: Whether the generalization is creative and meaningful (50\%).\\
* \textbf{Complexity}: Whether the problem has sufficient complexity and challenge (50\%).\\
\#\# [Final Requirements]\\
* Please ensure your generalization truly adds depth to the problem, not just changing surface parameters.\\
* Please ensure that the Prolog code has solutions in the end (not trivial solutions).}
\end{tcolorbox}

\section{Prolog as Prototype for Logic}
\label{appendix:pe_prolog}
\lstset{
  basicstyle=\footnotesize\ttfamily,
  breaklines=true,
  breakatwhitespace=false,
  columns=flexible,
  backgroundcolor=\color{gray!10},
  frame=single,
  framerule=0.5pt,
  xleftmargin=3pt,
  xrightmargin=3pt,
  aboveskip=6pt,
  belowskip=6pt,
  captionpos=b
}

\begin{table}[!tb]
\centering
\small
\renewcommand{\arraystretch}{1.2}
\begin{tabular}{@{}p{0.28\textwidth}p{0.28\textwidth}p{0.36\textwidth}@{}}
\toprule
\textbf{Knowledge Base} & \textbf{Query} & \textbf{Natural Language Description} \\
\midrule
\vspace{1mm}
\begin{lstlisting}[frame=single]
% Facts about parent relationships
parent(john, bob).
parent(john, lisa).
parent(mary, bob).
parent(mary, lisa).
parent(bob, ann).
parent(bob, james).
parent(lisa, carol).

% Rules defining relationships
father(X, Y) :- 
    parent(X, Y), 
    male(X).

mother(X, Y) :- 
    parent(X, Y), 
    female(X).

grandparent(X, Z) :- 
    parent(X, Y), 
    parent(Y, Z).

sibling(X, Y) :- 
    parent(P, X), 
    parent(P, Y), 
    X \= Y.

% Facts about gender
male(john).
male(bob).
male(james).
female(mary).
female(lisa).
female(ann).
female(carol).
\end{lstlisting} &
\vspace{1mm}
\begin{lstlisting}[frame=single]
% Who are Bob's children?
?- parent(bob, Child).
Child = ann ;
Child = james.

% Who are the grandparents of Carol?
?- grandparent(GP, carol).
GP = john ;
GP = mary.

% Are Lisa and Bob siblings?
?- sibling(lisa, bob).
true.

% Find all siblings
?- sibling(X, Y).
X = bob, Y = lisa ;
X = lisa, Y = bob ;
X = ann, Y = james ;
X = james, Y = ann.

% Is Mary a grandmother?
?- mother(mary, P), parent(P, _).
P = bob ;
P = lisa.
\end{lstlisting} & 
\vspace{1mm}
\setlength{\parskip}{0.5em}
"We have a family with parents, children, and various family relationships. John and Mary have two children, Bob and Lisa. Bob has two children, Ann and James, while Lisa has one child, Carol. We want to represent these relationships and query various family connections, such as identifying grandparents, siblings, and other family relations."

This classic family relationship problem is formalized through Prolog facts and rules, which translate natural language relationships into logical predicates and inference rules. The knowledge base contains both explicit facts (direct parent relationships, gender) and rules that define derived relationships (grandparent, sibling). Prolog's inference engine then answers queries by applying logical resolution to find all possible solutions that satisfy the given logical conditions. \\
\bottomrule
\end{tabular}
\caption{Prolog knowledge base, queries and their description}
\label{tab:prolog_examples}
\end{table}
\label{appendix:prolog_intro}

Prolog (Programming in Logic)~\citep{covington1994prolog} is a declarative programming language based on first-order predicate logic. As a logic prototype, Prolog provides an elegant framework that represents logical reasoning through a fact and rule-based system, enabling direct modeling of relational knowledge and inference processes.

\textbf{Structure and Components}: The fundamental structure of Prolog is built around a knowledge base composed of facts and rules, formally represented as a program $P = F \cup R$. Here $F = \{f_1, f_2, \ldots, f_m\}$ represents the set of facts, where each fact $f_i$ is a predicate consisting of a relation name and terms. Meanwhile, $R = \{r_1, r_2, \ldots, r_n\}$ represents the set of rules, where each rule $r_j$ is formulated as $H :- B_1, B_2, \ldots, B_k$, with $H$ being the head (conclusion) and the conjunction of $B_i$ forming the body (conditions).

\textbf{Logic Representation}: The core of Prolog's reasoning power lies in its unification algorithm and resolution strategy. Predicates represent relations between objects, variables allow for expressing general patterns, and rules enable the definition of complex logical relationships and inferences. This representation allows Prolog to model logical entailment, recursive definitions, and complex queries with remarkable clarity. Prolog's evolution has incorporated extensions such as constraint logic programming, tabling, and probabilistic logic programming, enhancing its expressive capabilities for diverse reasoning tasks.

\textbf{Cognitive Alignment}: Prolog's expression paradigm closely aligns with human cognitive processes in reasoning tasks, particularly in applying deductive inference, pattern matching, recursive thinking, and relational reasoning. This alignment makes Prolog an ideal prototype for capturing the structure of logical reasoning problems in a way that mirrors human thought processes. The declarative nature of Prolog allows for focusing on the logical relationships (knowledge representation) rather than procedural details, creating a representation that reflects the abstract reasoning patterns used in human logical thinking.

\section{PDDL as Prototype for Planning}
\label{appendix:pddl_intro}

PDDL (Planning Domain Definition Language)~\citep{aeronautiques1998pddl} is a standardized formal language designed to represent automated planning problems. As a logic prototype, PDDL provides a rich, expressive framework that captures the essential elements of planning domains through a declarative specification approach.

\textbf{Structure and Components}: The fundamental structure of PDDL emphasizes a clear separation between domain knowledge and problem-specific details, represented formally as a tuple $(D, P)$. Here $D = (T, P, A)$ represents the domain definition, containing a type hierarchy $T$ (organizing objects into categories), a set of predicates $P$ (defining relations between objects), and a collection of action schemas $A$ (specifying state transitions). Meanwhile, $P = (O, I, G)$ represents the problem instance, containing a set of objects $O$ (populating the domain), an initial state $I$ (describing the starting configuration), and a goal specification $G$ (defining desired conditions).

\textbf{Action Representation}: The core of PDDL's expressive power lies in its action representation. Each action schema $a \in A$ is defined as a tuple $(\textit{params}, \textit{pre}, \textit{eff})$, where $\textit{params}$ are typed variables representing action parameters, $\textit{pre}$ is a logical formula specifying preconditions, and $\textit{eff}$ describes effects as additions and deletions to the state. This representation enables PDDL to model causality, constraints, and state evolution with high fidelity to human planning processes. PDDL's evolution through various extensions (PDDL2.1, PDDL2.2, PDDL3.0, etc.) has progressively incorporated features such as numeric fluents, durative actions, preferences, and trajectory constraints, enhancing its representational capabilities.

\textbf{Cognitive Alignment}: PDDL's expression paradigm closely aligns with human cognitive processes in planning tasks, particularly in analyzing preconditions before action execution, predicting changes resulting from actions, decomposing complex goals into achievable subgoals, and reasoning about action sequences and their consequences. This alignment makes PDDL an ideal prototype for capturing the logical structure of planning problems in a way that facilitates both computational implementation and human understanding. The declarative nature of PDDL allows for focusing on the "what" (problem definition) rather than the "how" (solution procedure), creating a representation that mirrors abstract reasoning patterns used in human planning.


\lstset{
  basicstyle=\footnotesize\ttfamily,
  breaklines=true,
  breakatwhitespace=false,
  columns=flexible,
  backgroundcolor=\color{gray!10},
  frame=single,
  framerule=0.5pt,
  xleftmargin=3pt,
  xrightmargin=3pt,
  aboveskip=6pt,
  belowskip=6pt,
  captionpos=b
}

\begin{table}[!tb]
\centering
\small
\renewcommand{\arraystretch}{1.2}
\begin{tabular}{@{}p{0.28\textwidth}p{0.28\textwidth}p{0.36\textwidth}@{}}
\toprule
\textbf{Domain} & \textbf{Problem} & \textbf{Natural Language Description} \\
\midrule
\vspace{1mm}
\begin{lstlisting}[frame=single]
(define (domain n-puzzle-typed)
  (:requirements :typing)
  (:types position tile)
  (:predicates
    (at ?tile - tile ?position - position)
    (neighbor ?p1 - position ?p2 - position)
    (empty ?position - position))

  (:action move
    :parameters (?tile - tile 
                 ?from ?to - position)
    :precondition 
      (and (neighbor ?from ?to)
           (at ?tile ?from)
           (empty ?to))
    :effect 
      (and (at ?tile ?to) 
           (empty ?from)
           (not (at ?tile ?from)) 
           (not (empty ?to)))))
\end{lstlisting} &
\vspace{1mm}
\begin{lstlisting}[frame=single]
(define (problem n-puzzle-3)
  (:domain n-puzzle-typed)
  (:objects 
    p_1_1 p_1_2 p_1_3 
    p_2_1 p_2_2 p_2_3 
    p_3_1 p_3_2 p_3_3 - position 
    t_1 t_2 t_3 t_4 t_5 t_6 t_7 t_8 - tile)
  (:init
    (at t_2 p_1_1)
    (at t_3 p_1_2)
    (at t_7 p_1_3)
    (at t_5 p_2_1)
    (at t_4 p_2_2)
    (empty p_2_3)
    (at t_8 p_3_1)
    (at t_6 p_3_2)
    (at t_1 p_3_3)
    (neighbor p_1_1 p_1_2)
    (neighbor p_1_2 p_1_1)
    )
  (:goal (and
    (at t_1 p_1_1)
    (at t_2 p_1_2)
    (at t_3 p_1_3)
    (at t_4 p_2_1)
    (at t_5 p_2_2)
    (at t_6 p_2_3)
    (at t_7 p_3_1)
    (at t_8 p_3_2))))
\end{lstlisting} & 
\vspace{1mm}
\setlength{\parskip}{0.5em}
"You have a 3×3 grid with eight numbered tiles and one empty space. Tiles can slide into the adjacent empty space. The tiles begin in a scrambled configuration, and you must rearrange them to have tiles 1-8 in numerical order."

This classic puzzle game, commonly referred to as the "N-puzzle" in natural language, is formalized through PDDL domain and problem definitions, which essentially translate the natural language problem into the PDDL code space. The actual feasible planning solution is subsequently derived by a solver. \\
\bottomrule
\end{tabular}
\caption{PDDL domain, problem and their description}
\label{tab:pddl_examples}
\end{table}

\end{document}